\pdfoutput = 1

\documentclass[letterpaper, 10 pt, conference]{ieeeconf}  

\usepackage{cite}
\usepackage{amsmath,amssymb,amsfonts}
\usepackage{graphicx}
\usepackage{textcomp}
\usepackage{xcolor}                                   
\usepackage{kotex}      
\usepackage{algpseudocode}
\usepackage{algorithm}
\usepackage{setspace}
\usepackage{multirow}
\usepackage{makecell}
\usepackage{easyReview}
\usepackage{tablefootnote}
\usepackage{hyperref}
\usepackage{wrapfig}
\usepackage{wrapfig,lipsum,booktabs}
\let\labelindent\relax
\usepackage{enumitem}
\usepackage{bm}
\usepackage{lipsum}
\usepackage[T1]{fontenc}
\usepackage{array}
\usepackage{makecell}

\IEEEoverridecommandlockouts                              

\title{\LARGE \bf
A Learning Framework for Diverse Legged Robot Locomotion \\ Using Barrier-Based Style Rewards}

\author{Gijeong Kim$^{1*}$, Yong-Hoon Lee$^{1*}$, and Hae-Won Park$^{1\dagger}$}

\begin{document}
\maketitle
\def\thefootnote{}\footnotetext[0]{This research was supported in part by the Challengeable Future Defense Technology Research and Development Program through the Agency For Defense Development(ADD) funded by the Defense Acquisition Program Administration(DAPA) in 2024(No.912768601), and in part by the Technology Innovation Program(or Industrial Strategic Technology Development Program-Robot Industry Technology Development)(00427719, Dexterous and Agile Humanoid Robots for Industrial Applications) funded by the Ministry of  Trade Industry \& Energy(MOTIE, Korea).}
\def\thefootnote{1}\footnotetext[0]{Gijeong Kim, Yong-Hoon Lee, and Hae-Won Park are with the Department of Mechanical Engineering, Korea Advanced Institute of Science and Technology, Yuseong-gu, Daejeon 34141, Republic of Korea. {\tt\small haewonpark@kaist.ac.kr}}
\def\thefootnote{*}\footnotetext[0]{These authors contributed equally to this work}
\def\thefootnote{$\dagger$}\footnotetext[0]{Corresponding author}
\def\thefootnote{}\footnotetext[0]{Supplementary video: \href{https://youtu.be/JV2_HfTlOKI}{https://youtu.be/JV2\_HfTlOKI}}

\begin{abstract}
This work introduces a model-free reinforcement learning framework that enables various modes of motion (quadruped, tripod, or biped) and diverse tasks for legged robot locomotion.
We employ a motion-style reward based on a relaxed logarithmic barrier function as a soft constraint, to bias the learning process toward the desired motion style, such as gait, foot clearance, joint position, or body height.
The predefined gait cycle is encoded in a flexible manner, facilitating gait adjustments throughout the learning process. 
Extensive experiments demonstrate that \textit{KAIST HOUND}, a 45\,kg robotic system, can achieve biped, tripod, and quadruped locomotion using the proposed framework; quadrupedal capabilities include traversing uneven terrain, galloping at 4.67\,m/s, and overcoming obstacles up to 58\,cm (67\,cm for \textit{HOUND2}); bipedal capabilities include running at 3.6\,m/s, carrying a 7.5\,kg object, and ascending stairs-all performed without exteroceptive input. 
\end{abstract}

\vspace{-0.3cm}
\section{INTRODUCTION}
Controlling legged robots is a challenging task due to their inherent complexity, including hybrid and unstable dynamics~\cite{wensing2023optimization,ha2024learning}. To achieve animal-like natural motion, key locomotion characteristics such as foot clearance, body height, and gait preferences must be specified. In most traditional model-based approaches, these characteristics are typically defined by control designers, separating the generation of natural motion from the control problem~\cite{cheetah3,Donghyun,hong2020real}. These methods often rely on foothold position heuristics~\cite{raibert1984experiments} and manually tuned swing leg trajectories with predefined gaits. However, even with natural gait motion, these predefined elements struggle to adapt to environmental changes, posing challenges in responding to uncertainties in unmodeled components.
Alternatively, reinforcement learning (RL) employs various reward terms, such as regularization for base movement, joint positions, orientations, and velocity tracking \cite{ha2024learning,hwangbo2019learning,lee2020learning}, which are manually tuned through trial and error to achieve natural motion and robust control. While this approach can yield strong control performance, locomotion becomes sensitive to the specific weights and functions chosen for each reward, potentially affecting training outcomes and leading to inconsistent results.

To achieve desirable gait motion while mitigating the challenge of complex reward tuning in RL, several approaches have been explored.
By leveraging prior motion data from animal motion capture or trajectory optimization, imitation reward terms~\cite{escontrela2022adversarial,wu2023learning,fuchioka2023opt} or frameworks~\cite{youm2023imitating} incorporated in RL can realize targeted motions.
Alternatively, without prior motion data, using gait-specific rewards can enhance the versatility of locomotion (walking, running, hopping, skipping) in biped robots with a predefined preferred gait~\cite{siekmann2021sim}. Recently, Kim et al.~\cite{kim2024not} demonstrated that applying constraints in constrained RL to define operational motion ranges reduces reward engineering efforts and enhances the framework’s generalizability across different robot types.

Here, we introduce a learning framework that explicitly guides motion style with a preferred gait, rather than strictly constraining specific motion characteristics or gait sequences, to enhance task versatility for legged robots while facilitating animal-like natural motion. To achieve this flexible guidance, we propose to utilize a reward function based on the relaxed-logarithmic barrier function, a soft constraint type originally used in trajectory optimization~\cite{hauser2006barrier}. Unlike constrained RL approaches that prioritize safe learning and strict constraint satisfaction~\cite{lee2023evaluation}, our method focuses on guiding motion characteristics while maintaining flexibility. The preferred gait is also shaped by our proposed reward function, which features an explicitly tunable margin that facilitates adjustments of stance and swing times based on task requirements.


We demonstrate the scalability of the proposed framework across various tasks, including hiking, overcoming obstacles up to 67\,cm (nearly 0.96 times the leg’s full extension), galloping at 4.67\,m/s, tripod walking, and executing bipedal maneuvers such as running at 3.6\,m/s and carrying a 7.5\,kg object, using quadruped robots HOUND and HOUND2~\cite{shin2022design}.

\begin{figure*}
    \begin{center}
        \includegraphics[width=1.0\linewidth]{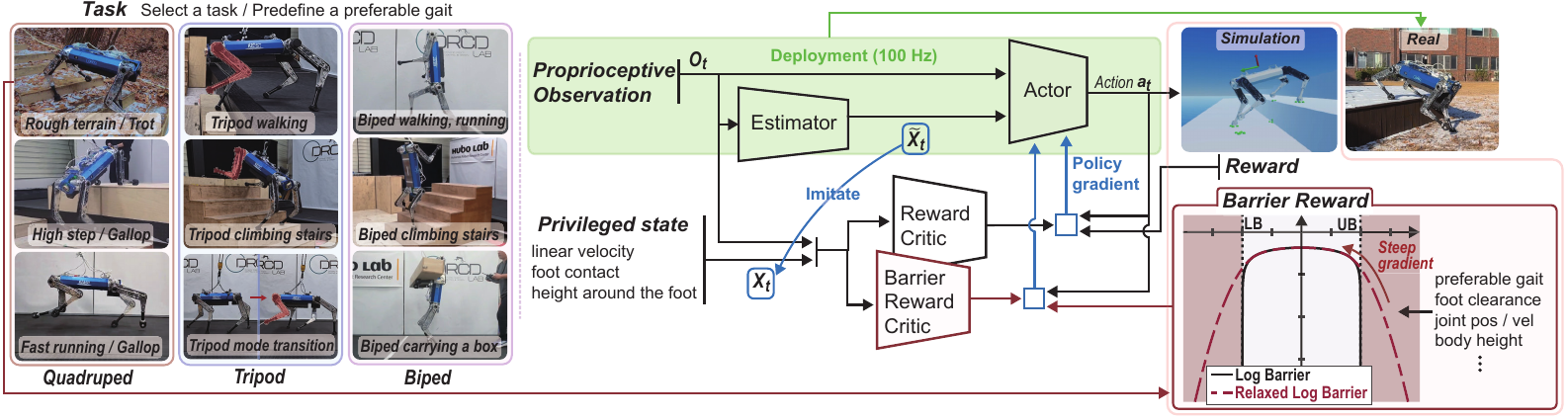}
    \end{center}
    \vspace{-0.22cm}
    \caption{Overview of the proposed RL framework. The policy learns to follow the commanded velocity with task-specific preferred gaits and motion styles. Within a single quadruped robot system, three different modes (quadruped, tripod, biped) are implemented, accommodating a variety of tasks: quadruped mode for traversing rough terrain, overcoming high steps, and fast running with a galloping gait; tripod mode for walking; and biped mode for walking, climbing stairs, and carrying a box. Task-specific gaits and styles are enforced via soft constraints in the barrier reward, which imposes a steep gradient in the constraint-violation region, thereby facilitating the learning of the desirable motion style.}
    \label{figure:overview}
    \vspace{-10pt}
\end{figure*}

\section{Related Works}
\textbf{Guiding to Desired Motion Style} \enspace 
To facilitate natural motion without complex reward engineering, previous studies have incorporated additional reward terms to define motion styles: 
If a prior dataset is available, discriminator-based style rewards~\cite{peng2021amp,escontrela2022adversarial,vollenweider2023advanced,wu2023learning} or direct reward formulations (e.g., reducing discrepancies between reference and resultant motion)~\cite{peng2018deepmimic,fuchioka2023opt} can be used to imitate prior motions. However, this approach can be limited for hard-to-obtain data, such as tripod locomotion or biped locomotion of quadruped animals. Alternatively, Kim et al.~\cite{kim2024not} explicitly define locomotion styles by imposing constraints on physical quantities, such as joint angles and velocities, in constrained RL. They employ Interior Point Optimization (IPO)\cite{liu2020ipo}, which manages constraints with logarithmic barrier functions. This requires additional algorithms to handle the infinity regions during constraint violations\cite{kim2024not,lee2023evaluation}. Our approach explicitly defines locomotion styles through constraints; however, unlike previous methods, we incorporate a relaxed logarithmic barrier~\cite{hauser2006barrier} into the reward function. This eliminates the need for additional algorithms to handle infinite log-function regions and simplifies implementation.

\textbf{Gait Encoding} \enspace 
In RL for legged robots, previous studies have explored optimal gait motions either without any bias~\cite{hwangbo2019learning} or with only implicit guidance, such as penalizing excessive swing times~\cite{ji2022concurrent,rudin2022learning}. However, to achieve diverse and natural gait motions, the preferred gait can be encoded in the action space~\cite{lee2020learning,bellegarda2022cpg} or modulated through rewards~\cite{wu2023learning,siekmann2021sim}. Based on Central Pattern Generators (CPGs), the gait is either guided through an encoded time phase~\cite{lee2020learning,miki2022learning} or discovered by setting the action as intrinsic parameters for the CPG~\cite{bellegarda2022cpg}. Alternatively, by utilizing AMP-based style rewards, prior motion data can guide desirable gait motions~\cite{lee2020learning,miki2022learning}. 
Even without prior data, gait-specific rewards designed to penalize inappropriate contact timings within the preferred gait settings can effectively enforce desirable gaits~\cite{siekmann2021sim}. 
Our work employs gait-specific rewards with increased flexibility, enforcing the gait within a tunable margin. This approach facilitates adjustments to stance and swing times, even enabling extended stance periods for tripod motion under standard trot gait settings in a quadruped.

\section{Methods}
\label{sec:method}
Across all tasks, including overcoming high obstacles, fast running, and traversing uneven terrain, the main objective of the controller is to track commanded velocities under blind settings, using only proprioceptive information without exteroceptive sensors. Fig.~\ref{figure:overview} provides an overview of our RL framework, which utilizes two types of rewards: the proposed barrier reward, $R_{\text{barrier}}$, and the standard regularization reward, $R_{\text{standard}}$. The barrier reward comprises soft constraints that define desired styles and impose sharp penalties for significant deviations from these styles, thereby guiding robot motion towards desirable behaviors. Based on the target task and motion mode (biped, tripod, quadruped), the preferred gaits and desired style for $R_{\text{barrier}}$ are determined. 


\subsection{Relaxed Logarithmic Barrier Reward}
\label{subsec:relaxedlog}
We implemented soft constraints within $R_{\text{barrier}}$ using a relaxed logarithmic barrier function; originally proposed to incorporate inequality constraints into the cost function for trajectory optimization~\cite{grandia2019feedback, hauser2006barrier}. This relaxed log-barrier function acts as a log-barrier (as in optimization~\cite{boyd2004convex}) within the interior feasible region and can be formulated as a quadratic function outside the constraint boundary. The transition occurs at a distance $\delta$ from the boundary. The relaxed barrier function $B(z;\delta)$ for the constraint $z \geq 0$ is as follows:
\begin{align}
B(z;\delta) =
  \begin{cases}
    \;\log{z}, & \quad z> \delta, \\
    \;\log{\delta}-\dfrac{1}{2}\left[\left(\dfrac{z-2\delta}{\delta}\right)^{2}-1\right], & \quad z\leq \delta. \nonumber
  \end{cases}    
\end{align} 
The main distinction from the log-barrier function is that the relaxed function $B(z;\delta)$ allows gradient steps under \textit{constraint-violation} ($z < 0$), whereas the log-barrier function cannot be evaluated in that region, yielding infinite values. 
Therefore, in contrast to previous work that requires additional algorithms to handle infinite values with log-barrier functions in constrained RL with IPO~\cite{kim2024not,lee2023evaluation}, our approach eliminates this need.
Additionally, as $\delta \rightarrow 0$, the relaxed function $B(z;\delta)$ becomes a steeper quadratic function in the region where $z\leq \delta$, reverting to the logarithmic function. Therefore, using this steep gradient information, the policy can be guided from the constraint-violation region to the constraint-satisfying region.

Define the $k$-th constraints by $d_k^{\text{lower}} \leq C_k(s_t, a_t, s_{t+1}) \leq d_k^{\text{upper}}$ for each $k \in \{1, \ldots, K\}$, where $C_k(s_t, a_t, s_{t+1})$ is a constraint variable that is a function of the state $s_t, s_{t+1}$ and action $a_t$, similar to the reward function. $d_k^{\text{lower}}$ and $d_k^{\text{upper}}$ are the lower and upper boundaries, respectively, and $K$ is the total number of constraints. Then, the barrier reward is formulated as a weighted sum of relaxed barrier functions for each constraint:
\begin{equation}
\label{eq:R_barrier}
\begin{aligned}
R_{\text{barrier}}=\sum_{k=1}^{K}\alpha_{k}\big[B&(-d_k^{\text{lower}}+C_{k}(s_{t}, a_{t}, s_{t+1});\delta_k) \\
 + &B(d_k^{\text{upper}}- C_{k}(s_{t}, a_{t}, s_{t+1});\delta_k)\big].
\end{aligned}
\end{equation}
where \(\alpha_k\), the weight for each barrier function, is equally set to 0.1 for all \(k \in \{1, \dots, K\}\).

In trajectory optimization, Hauser et al.~\cite{hauser2006barrier} iteratively reduce $\delta$ and $\alpha_k$ to ensure convergence to the original constrained optimization problem. In this work, our goal is to guide motion style rather than strictly enforce constraints. Therefore, we keep $\delta$ and $\alpha_k$ constant to simplify the training process, as we verified that this setting still effectively achieves the desired motion style.


\textbf{Multi-Critic} \enspace Inspired by previous work~\cite{kim2024not,zargarbashi2024robotkeyframing}, independent critics process the barrier reward $R_{\text{barrier}}$ and standard reward $R_{\text{standard}}$ exclusively, with each critic addressing different aspects of each reward~\cite{zargarbashi2024robotkeyframing}. Moreover, this approach prevents the policy from opting for early termination due to the sharp penalties applied by the barrier rewards; where early termination, designed to prevent contact with the environment except by the calf and feet, incurs a penalty on the standard reward when it occurs~\cite{ji2022concurrent}. 

\textbf{Policy Updates} \enspace Proximal Policy Optimization (PPO) \cite{schulman2017proximal} is employed for policy updates. The clipped surrogate loss is computed using the sum of two equally weighted advantage functions, one for each reward (barrier and standard). This constant weight is feasible because we normalize and standardize the advantage functions before computing the surrogate loss, as in~\cite{kim2024not, lee2023evaluation, zargarbashi2024robotkeyframing}; this ensures the relative magnitude of the barrier reward to the standard reward remains unchanged~\cite{lee2023evaluation,zargarbashi2024robotkeyframing}, even with sharp penalties from the barrier reward, thus facilitating a stable learning process.

\subsection{Gait Encoding and Foot Clearance}
\label{subsec:contact}
\begin{figure}
    \includegraphics[width=1.0\columnwidth]{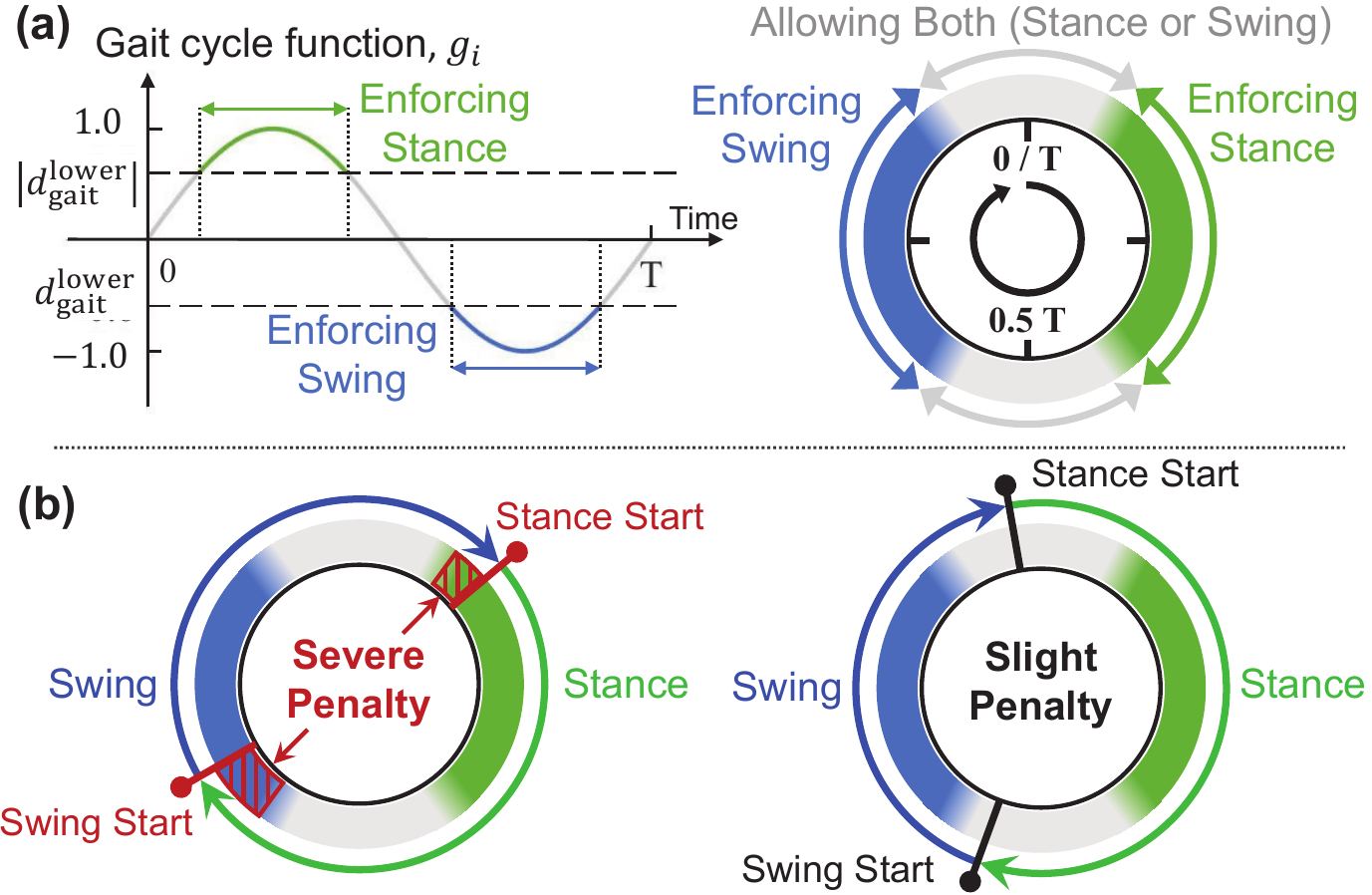}
  \caption{Graphical representation of gait cycle and desired gait enforcement: (a) illustrates the gait cycle function \(g_{i}(t)\) with no phase offset. The constraint boundary \(d_{\text{gait}}^{\text{lower}}\) is set to less than zero, enforcing stance or swing only when \(|g_{i}(t)| \geq |d_{\text{gait}}^{\text{lower}}|\); (b) indicates when the gait constraint (\( f_{i} \geq d_{\text{gait}}^{\text{lower}} \)) imposes a severe penalty (left) and a lesser penalty (right). Consequently, the starting points of stance and swing phases are adjusted within the grey region (right).}
  \label{figure:contact}
  \vspace{-15pt}
\end{figure}

We predefined preferred gaits for specific tasks. To inform the phase during gait motion, the gait cycle function \(g_{i}(t)\) is defined as a sine function; for the \(i\)-th foot, \(g_{i}(t)\) at time \(t\) with a period \(T\) is expressed as:
    $g_{i}(t) = \sin\left(2\pi(\frac{t}{T}+\phi_i)\right)$, where \(\phi_i\) is the phase offset for the \(i\)-th foot. The function \(g_{i}(t)\) determines the constraint variable for foot contact:
\begin{equation}
\begin{aligned} 
\label{eq:constraint_variable_gait}
f_{i} =
  \begin{cases}
    g_{i}(t), & \quad\text{$i$-th foot contacts}, \\
    -g_{i}(t), & \quad\text{$i$-th foot does not contact}, \\
  \end{cases}
\end{aligned}    
\end{equation}
where, $f_{i}\in[-1,1]$. Therefore, setting the constraint \( f_{i} \geq d_{\text{gait}}^{\text{lower}} \) with \(d_{\text{gait}}^{\text{lower}}=0\) strictly requires a stance for the first half of the gait period and a swing for the second half. To enable more adaptive gait patterns, we lower the constraint boundary \(d_{\text{gait}}^{\text{lower}}\); for instance, \(f_{i} \geq -0.6\) requires stance only when \(g_{i}(t) \geq 0.6\) and swing when \(g_{i}(t) \leq -0.6\) (Fig.~\ref{figure:contact} (a)). In other regions, either stance or swing satisfies \( f_{i} \geq -0.6 \), allowing the stance and swing times to vary from 30\% to 70\% of \(T\) without significant penalty (Fig.~\ref{figure:contact} (b)). Notably, this approach enforces the desired gait frequency while allowing flexible adaptation of stance and swing times.

Foot clearance is also formulated based on the gait cycle function \(g_{i}(t)\). The constraint variable for foot clearance, \(l_{i}\), is defined to lift the foot higher than the surrounding terrain with a margin, when swing is enforced ($g_{i}(t) \leq d_\text{gait}^{\text{lower}}$):
\begin{equation}
\begin{aligned} 
l_{i} =
  \begin{cases}
  \label{eq:constraint_variable_clearance}
    p_{i}-\big(\text{max}\left(H_{\text{sample},i}\right) + p_{\text{des}}\big), & \enspace g_{i}(t) \leq d_\text{gait}^{\text{lower}}, \\
    0, & \enspace g_{i}(t) > d_\text{gait}^{\text{lower}}, 
  \end{cases}
\end{aligned}        
\end{equation}
where \(p_{i}\) is the height of the \(i\)-th foot, \(H_{\text{sample},i}\) is the sampled terrain height around the \(i\)-th foot (within 5-cm), and \(p_{\text{des}}\) is the desired foot height (set to 0.15 m).
The margin for foot lift can be adjusted by the constraint boundary \(d_\text{clearance}^{\text{lower}}\) (e.g., set to -0.08), where \(l_{i} \geq d_\text{clearance}^{\text{lower}}\).

\textbf{Task-dependent Gait} \enspace Task-specific gaits are set by adjusting the gait period \(T\) and phase offsets \(\phi_i\) for each leg, denoted as RH, LH, RF, LF for \(i=1,2,3,4\), respectively. For the quadruped mode traversing rough terrain, we employ a trotting gait with \(T=0.72\,\text{sec}\), setting diagonal leg pairs \(\phi_{1,3}=0\) and \(\phi_{2,4}=0.5\). Based on this basic trot setting, tripod and biped modes are implemented: 1) In tripod mode, only the constraint variables \(f_{i'}\) (Eq.~\ref{eq:constraint_variable_gait}) and \(l_{i'}\) (Eq.~\ref{eq:constraint_variable_clearance}) are adjusted for the non-contacting \(i'\)-th foot. Here, \(f_{i'}\) is set to \(-1\) if contacting and \(1\) if not (for non-contact), while \(l_{i'}\) is set to always adhere to the first case in Eq.~\ref{eq:constraint_variable_clearance} (for continuous foot lift). 2) In biped mode, constraints for gait and foot clearance are enforced only for the two contacting feet.

In quadruped mode for overcoming high steps, we employ a transverse gallop gait with \(T=0.82\,\text{sec}\) and phase offsets \([0, 0.2, 0.5, 0.7]\) for \(\phi_{1}, \phi_{2}, \phi_{3}, \phi_{4}\) respectively. For fast running tasks, we use a rotary gallop gait with \(T=0.72\,\text{sec}\) and phase offsets \([0, 0.2, 0.7, 0.5]\). Using the heuristics from~\cite{park2017high}, for command velocities above 1.75\,m/s, stance time \(T_{\text{st}}\) is computed as \(L/v^{\text{cmd}}_x\) where \(L\) is the stride length (0.63\,m) and \(v^{\text{cmd}}_x\) is the forward velocity command. Swing time \(T_{\text{sw}}\) remains fixed at 0.36\,sec across all velocities. Phase offsets are set as \([0, 0.4\,\frac{T_{\text{st}}}{T}, 0.5 + 0.4\,\frac{T_{\text{st}}}{T}, 0.5]\), ensuring continuous adaptation at 1.75\,m/s.

\subsection{Mode-dependent Barrier Reward Design}
We utilize the barrier reward to define: 1) motion characteristics including preferred gait, foot clearance, body height, and joint position; 2) targeted tasks (velocity tracking); and 3) regularization terms for base motion and joint velocity. For each constraint in Eq.~\ref{eq:R_barrier}, we define lower and upper bounds (\(d^{\text{lower}}, d^{\text{upper}}\)) and adjust the gradient steepness in the constraint-violation region using \(\delta\). Barrier reward settings for quadruped and biped modes are listed in Table~\ref{tab:BarrierReward}. 

\begin{table}[htb!]
\caption{Barrier reward function.}
\label{tab:BarrierReward}
\scriptsize 
\setlength{\tabcolsep}{1.5pt}
\renewcommand{\arraystretch}{1.4}
\begin{tabular}{|cl|ccc|ccc|}
\hline
\multicolumn{2}{|c|}{\multirow{2}{*}{\textbf{Constraint Variable}}} & \multicolumn{3}{c|}{\textbf{Quadruped Task}} & \multicolumn{3}{c|}{\textbf{Biped Task}} \\ \cline{3-8} 
\multicolumn{2}{|c|}{}  & \multicolumn{1}{c|}{$d^{\text{lower}}$}  & \multicolumn{1}{c|}{$d^{\text{upper}}$}  & \textbf{$\delta$}
& \multicolumn{1}{c|}{$d^{\text{lower}}$}  & \multicolumn{1}{c|}{$d^{\text{upper}}$} & \textbf{$\delta$}             \\ \hline
\multicolumn{1}{|c|}{Gait}  & $f_i$   & \multicolumn{1}{c|}{$-0.6$}  & \multicolumn{1}{c|}{-$^\dagger$}  & $0.1$ & \multicolumn{1}{c|}{$-0.6$}  & \multicolumn{1}{c|}{-$^\dagger$}  & $0.1$ \\ \hline
\multicolumn{1}{|c|}{Foot clearance}  & $l_i[m]$  & \multicolumn{1}{c|}{$-0.08$}  & \multicolumn{1}{c|}{-$^\dagger$}  & $0.01$  & \multicolumn{1}{c|}{$-0.08$}  & \multicolumn{1}{c|}{-$^\dagger$}  & $0.01$  \\ \hline
\multicolumn{1}{|c|}{\multirow{3}{*}{Joint position}} 
& $q_{\text{roll},i}-q^{\text{nom}}_{\text{roll},i}[rad]$ & \multicolumn{1}{c|}{$-\pi/6$} & \multicolumn{1}{c|}{$\pi/6$} & \multirow{3}{*}{$0.08$} & \multicolumn{1}{c|}{$-\pi/7$} & \multicolumn{1}{c|}{$\pi/7$}  & \multirow{3}{*}{$0.08$} \\ \cline{2-4} \cline{6-7} 
\multicolumn{1}{|c|}{}  & $q_{\text{hip},i}-q^{\text{nom}}_{\text{hip},i}[rad]$ & \multicolumn{1}{c|}{$-\pi/4$}  & \multicolumn{1}{c|}{$\pi/4$}  &  & \multicolumn{1}{c|}{$-2\pi/5$}  & \multicolumn{1}{c|}{$2\pi/5$}  & \\ \cline{2-4} \cline{6-7}
\multicolumn{1}{|c|}{}  & $q_{\text{knee},i}-q^{\text{nom}}_{\text{knee},i}[rad]$   & \multicolumn{1}{c|}{$-2\pi/5$}  & \multicolumn{1}{c|}{$\pi/4$}  &  & \multicolumn{1}{c|}{$-\pi/5$}  & \multicolumn{1}{c|}{$\pi/5$}  &  \\ \hline
\multicolumn{1}{|c|}{\multirow{2}{*}{Body height}}  & ${_b}h_{F}[m]$  & \multicolumn{1}{c|}{\multirow{2}{*}{$0.52$}} & \multicolumn{1}{c|}{\multirow{2}{*}{$0.66$}} & \multirow{2}{*}{$0.04$} & \multicolumn{1}{c|}{$0.35$}  & \multicolumn{1}{c|}{$0.85$}  & \multirow{2}{*}{$0.03$} \\ \cline{2-2} \cline{6-7}
\multicolumn{1}{|c|}{}  & ${_b}h_{H}[m]$   & \multicolumn{1}{c|}{}  & \multicolumn{1}{c|}{}  &  & \multicolumn{1}{c|}{$1.05$}  & \multicolumn{1}{c|}{$1.55$}  &  \\ \hline
\multicolumn{1}{|c|}{\multirow{3}{*}{Target velocity}} & $v^{\text{cmd}}_{x}-v_{x}[m/s]$   & \multicolumn{1}{c|}{\multirow{3}{*}{$-0.4$}} & \multicolumn{1}{c|}{\multirow{3}{*}{$0.4$}}  & \multirow{3}{*}{$0.2$}  & \multicolumn{1}{c|}{\multirow{3}{*}{$-0.$4}} & \multicolumn{1}{c|}{\multirow{3}{*}{$0.4$}} & \multirow{3}{*}{$0.2$}  \\ \cline{2-2}
\multicolumn{1}{|c|}{} & $v^{\text{cmd}}_{y}-v_{y}[m/s]$   & \multicolumn{1}{c|}{}  & \multicolumn{1}{c|}{}  &  & \multicolumn{1}{c|}{}  & \multicolumn{1}{c|}{}  &  \\ \cline{2-2}
\multicolumn{1}{|c|}{}  &  $\omega^{\text{cmd}}_{z}-\omega_{z}[rad/s]$  & \multicolumn{1}{c|}{}  & \multicolumn{1}{c|}{}  &  & \multicolumn{1}{c|}{}  & \multicolumn{1}{c|}{}  &  \\ \hline
\multicolumn{1}{|c|}{\multirow{3}{*}{Base motion}}  & $\omega_x[rad/s]$   & \multicolumn{1}{c|}{\multirow{3}{*}{$-0.3$}} & \multicolumn{1}{c|}{\multirow{3}{*}{$0.3$}}  & \multirow{2}{*}{$0.3$}  & \multicolumn{1}{c|}{\multirow{3}{*}{$-0.3$}} & \multicolumn{1}{c|}{\multirow{3}{*}{$0.3$}} & \multirow{2}{*}{$0.3$}  \\ \cline{2-2}
\multicolumn{1}{|c|}{}  & $\omega_y[rad/s]$   & \multicolumn{1}{c|}{}  & \multicolumn{1}{c|}{}  &  & \multicolumn{1}{c|}{}  & \multicolumn{1}{c|}{}  &  \\ \cline{2-2} \cline{5-5} \cline{8-8} 
\multicolumn{1}{|c|}{}  & $v_z[m/s]$   & \multicolumn{1}{c|}{}  & \multicolumn{1}{c|}{}  & $0.2$  & \multicolumn{1}{c|}{}  & \multicolumn{1}{c|}{}  & $0.2$  \\ \hline
\multicolumn{1}{|c|}{Joint velocity}  & $\dot{q_j}[rad/s]$   & \multicolumn{1}{c|}{$-8$}  & \multicolumn{1}{c|}{$8$}  & $2.0$  & \multicolumn{1}{c|}{$-12$}  & \multicolumn{1}{c|}{$12$}  & $2.0$ \\ \hline
\end{tabular}
\vspace{0.1cm}

\footnotesize{$(\cdot)^{nom}$ and $(\cdot)^{cmd}$ denote nominal and commanded values, respectively. $(\cdot)_{i}$ refers to the $i$-th leg, and $(\cdot)^{j}$ to the $j$-th joint. $x$, $y$, and $z$ are defined in the body frame. $f$, $l$, $q$, $\dot{q}$, $v$, and $\omega$ represent gait constraint variable, foot clearance, joint position, joint velocity, body linear velocity, and body rotational velocity, respectively. ${_b}h_{F}$ and ${_b}h_{H}$ denote the heights of the front and hind roll joints from the ground, respectively, in the body frame. {$^\dagger$}For \(f_i\) and \(l_i\), the upper bounds are unnecessary, thus set to non-reachable values (2.0 and 1.0, respectively).}
\vspace{-5pt}
\end{table}

Each constraint variable in Table~\ref{tab:BarrierReward}, except for gait, corresponds to a physical quantity; consequently, lower and upper bounds are established to define the desirable operational region (related to \cite{kim2024not},\cite{lee2023evaluation},\cite{ranjan2024barrier}). For gait $f_i$, the lower bound is designed to facilitate adjustments in stance and swing times. Joint position and body height are constrained to operate near the nominal posture for each motion mode. Body height constraints are applied separately to the robot's front and hind sides to align with stair or slope inclinations in quadruped mode and to support an upright two-legged stance in biped mode. The target velocity constraint guides towards the desired command velocity, allowing deviations based on environmental challenges (e.g., overcoming high steps). Base motion and joint velocity constraints are set to prevent aggressive motions. 

\begin{figure}
\begin{center}
\includegraphics[width=1.0\columnwidth]{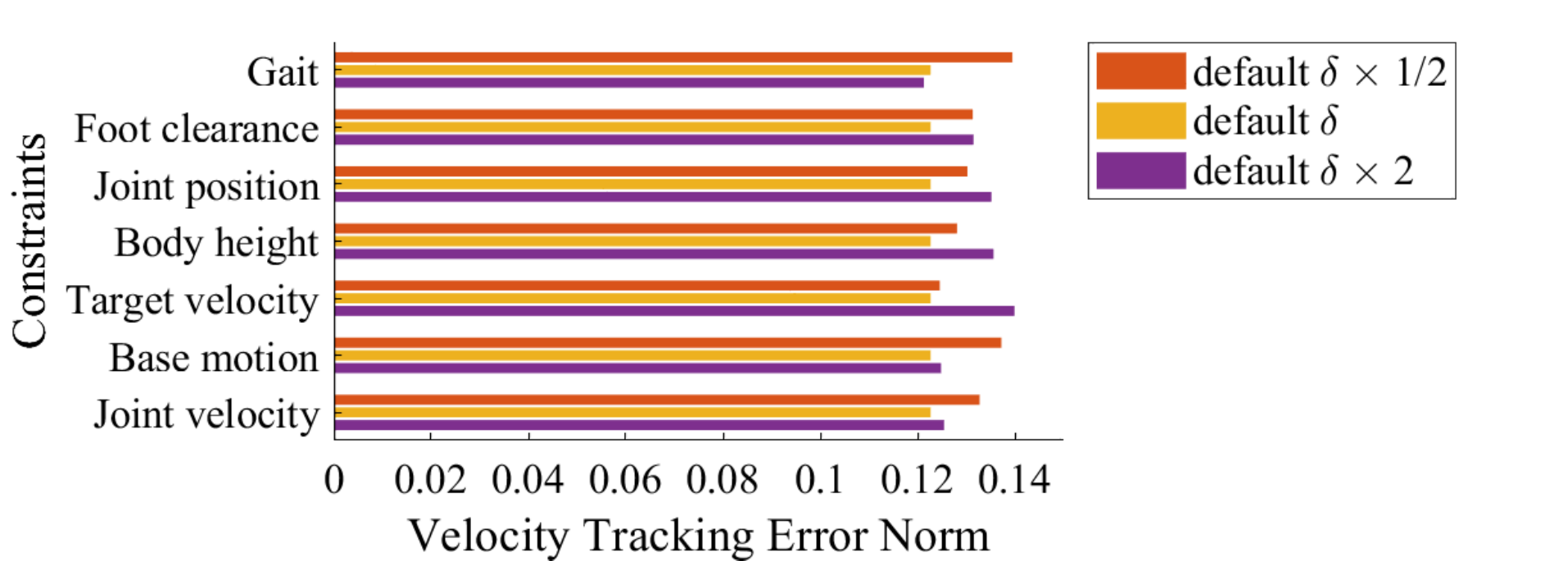}
\end{center}
\vspace{-10pt}
\caption{Velocity tracking performance for different \( \delta \) values (\( 1/2 \), 1, and 2 times the default setting in Table~\ref{tab:BarrierReward}) across constraints. Each configuration was evaluated in simulation under randomized conditions identical to the training setup for \textbf{Traversing Outdoor Terrain} in \ref{subsec:experiment}, where environments ranged from flat ground to challenging steps and slopes. The graph presents the mean tracking error norms of \( v_x \), \( v_y \), and \( \omega_z \) over a 5-second period across 2000 environments with randomized command velocities.}
\label{figure:sensitivity}
\vspace{-15pt}
\end{figure}

Within these predefined ranges, the parameter \( \delta \) determines the severity of penalties in constraint-violation regions; smaller values yield steeper gradients. This setting should align with each constraint's range; for example, we set \( \delta \) to 10–30\,\% of each constraint's range for most constraints, using a tighter 4\,\% for joint position to prevent singular poses and a wider 50\,\% for base motion to avoid overly sharp penalties that could hinder movement. 

The constraints enable task-specific adjustments. For instance, in high step climbing, the body height constraints are widened to allow a steeper body pitch for climbing maneuvers; in fast running with galloping, the lower boundary for gait is set to $-0.3$ for stricter enforcement, joint velocity boundaries are adjusted to $\pm16$ rad/s, and pitch velocity boundaries are expanded for command speed over $3.0$ m/s.

\textbf{Sensitivity of \( \delta \) selection} \enspace Fig.~\ref{figure:sensitivity} presents the velocity tracking performance for different \( \delta \) values across constraints. The results indicate that doubling or halving the default \( \delta \) in Table~\ref{tab:BarrierReward} did not lead to noticeable performance differences, suggesting that the proposed method is not too sensitive to \( \delta \) selection and can alleviate hyperparameter tuning. However, excessively small or large \( \delta \) can cause deviations from the desired motion (e.g., overly strict foot clearance enforcement leading to exaggerated foot lifting).


\begin{figure*}
    \centering
    \includegraphics[width=0.74\linewidth]{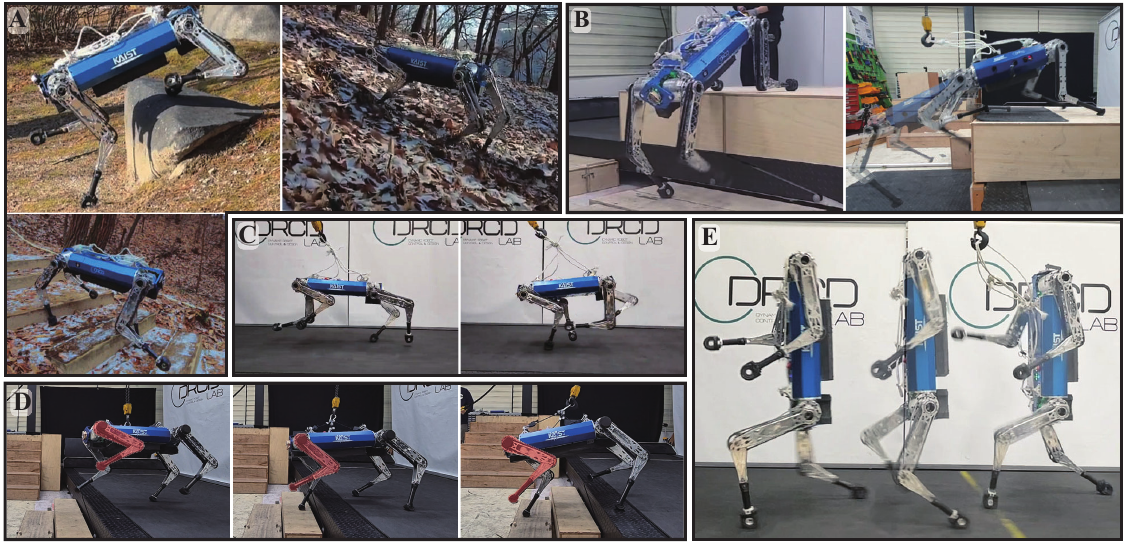}
    \caption{We develop a learning framework for challenging tasks for quadruped robots. The tasks include (A) locomotion over rough terrain, (B) overcoming a high step (58\,cm, 67\,cm), (C) agile running using a gallop gait, (D) tripod locomotion, and (E) executing humanoid walking motions.}
    \label{figure:task}
    \vspace{-15pt}
\end{figure*}

\subsection{Implementation Detail}
\label{section:Implementation}
\textbf{Learning Framework} \enspace As shown in Fig.~\ref{figure:overview}, we utilize a concurrent learning framework~\cite{ji2022concurrent} with an asymmetric actor-critic architecture~\cite{nahrendra2023dreamwaq, pinto2017asymmetric}, to leverage privileged information from simulations~\cite{lee2020learning,miki2022learning}. During actor network training, the estimator network concurrently learns to estimate the privileged state $\mathbf{x}_t$~\cite{ji2022concurrent}, comprising body’s linear velocity, foot contact state, and terrain information around the feet. This privileged state $\mathbf{x}_t$ and observation $\mathbf{o}_t$ feed into the critic network~\cite{nahrendra2023dreamwaq}, while the actor network uses the estimated state $\tilde{\mathbf{x}}_t$ and $\mathbf{o}_t$. The actor's action $\mathbf{a}_t$ set desired joint positions. Both actor and critic networks use a Multi-Layer Perceptron (MLP) structure of [256x128x64], and the estimator network uses a [256x128] MLP.

\textbf{Observation} \enspace 
The observation \(\mathbf{o}_t\) includes proprioceptive information such as body orientation, body angular velocity, joint positions and velocities, history of joint position errors and joint velocities, and relative foot positions in the body frame~\cite{ji2022concurrent}. It also includes previous actions, commanded velocity, cyclic functions, and a stand-mode indicator. The cyclic functions consist of $\sin\left(2\pi(\frac{t}{T})\right)$ and $\cos\left(2\pi(\frac{t}{T})\right)$ to indicate the gait-cycle phase. A stand-mode indicator triggers a standing-still motion~\cite{lee2020learning} when the norm of commanded velocity is less than 0.2. In stand-mode, barrier rewards for gait and foot clearance are adjusted to keep all legs in contact, and the cyclic functions are set to zero.

\textbf{Standard Reward} \enspace The purpose of the standard reward is additional regularization to facilitate sim-to-real transfer. Specifically, \(R_{\text{standard}}=r_{pos} \cdot \exp(0.2~r_{neg})\), following~\cite{ji2022concurrent} with modifications.
For the positive reward \(r_{pos}\), we use terms for tracking body linear and angular velocity commands, while for the negative reward \(r_{neg}\), we adopt regularization terms for foot slip, joint torque, action smoothness, and orientation. Please refer to previous work~\cite{ji2022concurrent} for detailed reward settings. Foot position term is added to \(r_{neg}\) to regulate the motion of foot in Cartesian coordinates relative to the nominal posture. In quadruped mode, the orientation reward is omitted, and an additional reward is introduced to penalize height differences between the front and hind sides, which helps avoid penalizing pitch tilt during ascents or descents on steep inclines or steps.

\textbf{Biped Mode Setting} \enspace For biped mode, the learning framework used in quadruped mode is applied with minor modifications: The nominal body orientation (identity matrix) and nominal posture are set to the upright bipedal stance pose. Therefore, the body orientation is computed by multiplying the measured orientation from the quadruped robot with a $-90^\circ$ pitch rotation matrix.
Due to the quadruped robot's point-foot design~\cite{shin2022design}, maintaining a standing-still motion on two legs is difficult. Therefore, we disable stand-mode and enforce continuous cyclic foot movement in biped mode.
\section{Results}
\label{sec:results}
\textbf{Training and Deployment Setup} \enspace We utilized RaiSim simulator for training~\cite{hwangbo2018per}. The training was conducted on an AMD Ryzen Threadripper PRO 5995WX and a single NVIDIA GeForce RTX 3080 Ti for 5 hours, over 10,000 iterations. During the learning process, 400 environments collected data at 100\,Hz in 4-second episodes, resulting in a batch size of 160,000. The control policy was deployed on the quadruped robots HOUND~\cite{shin2022design} and HOUND2.

\subsection{Comparison with the Baseline}
\textbf{Baseline} \enspace The baseline does not include a barrier reward or predefined gait, but employs the standard reward incorporating all reward types used in previous work~\cite{ji2022concurrent}. To ensure a fair comparison, we adjusted the gait-related reward (airtime reward in~\cite{ji2022concurrent}) to align with the preferred stance and swing times in our framework.

\begin{figure}
\begin{center}
\includegraphics[width=1.0\columnwidth]{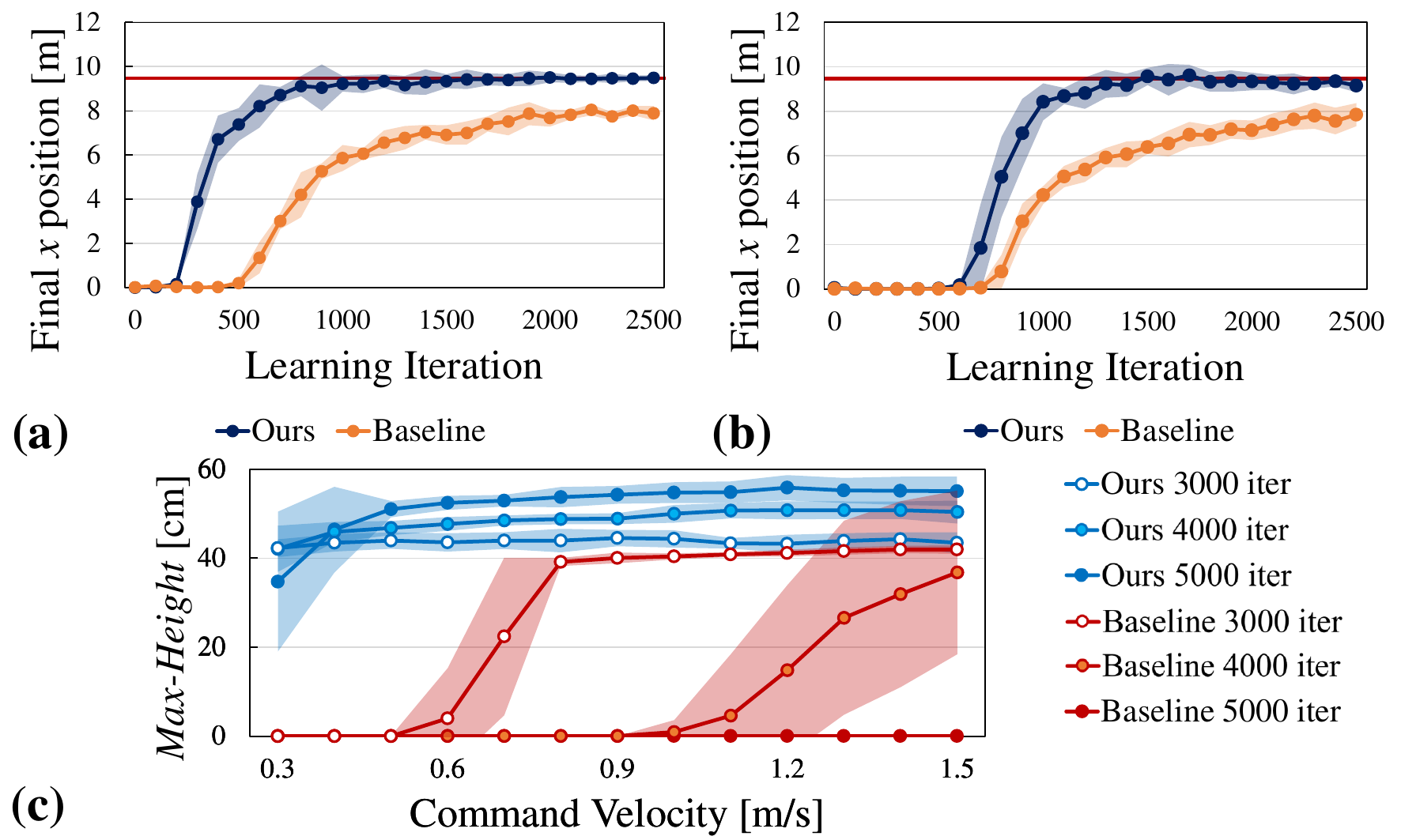}
\end{center}
\vspace{-10pt}
\caption{ We compare the performance of our method against the baseline in
(a) Quadruped locomotion on flat terrain
(b) Biped locomotion on flat terrain
(c) Quadruped locomotion for overcoming high steps. 10 policies trained with different random seeds, were evaluated for both our method and the baseline across 100 random initial conditions for each policy. Each graph shows the mean and standard deviation. In (a) and (b), both our method and the baseline converge with no further improvement after 2500 iterations.
}
\label{figure:comparison}
\vspace{-15pt}
\end{figure}

\textbf{Comparison 1: Learning to Walk} \enspace The training was conducted on flat terrain, and the robot learned to track random velocity commands. During evaluation, the robot aims to accelerate for one second and then maintain a speed of 1\,m/s, targeting a position as indicated by the red lines in Fig.~\ref{figure:comparison}. Fig.~\ref{figure:comparison}(a) and (b) represent the robot's final position along the \textit{x}-axis in quadruped and biped modes, respectively. Our framework achieves the target in fewer iterations than the baseline by guiding desirable gait motion and command velocity tracking from the early stages of learning, facilitated by the barrier reward. In contrast, the baseline gradually learns to stand, begins walking, and converges but does not fully reach the target position. 


\textbf{Comparison 2: Climbing High Step} \enspace For the high-step task, curriculum learning gradually increases step heights to 69\,cm over 7,600 iterations in a consecutive-step environment. Fig.~\ref{figure:comparison}(c) compares the step-climbing performance of our framework to that of the baseline. \textit{Max-Height} is defined as the highest step height at which the success rate of climbing three steps within the time limit exceeds 50\,\%. 
At 3,000 iterations, our framework consistently overcame the highest obstacles in the training environment for commands ranging from 0.3\,m/s to 1.5\,m/s. In contrast, the baseline failed to overcome obstacles at low-speed commands. As iterations progressed, our framework increased \textit{Max-Height} across most commands, aided by our explicit enforcement of foot clearance within the barrier reward. In contrast, the baseline exhibited performance degradation due to unstable foot clearance patterns, beginning with low speeds and ultimately leading to a stationary policy. 

\subsection{Real-World Experiments}
\label{subsec:experiment}
We validated our learning framework through real-world experiments with quadruped, tripod, and biped modes tasks.
Our framework utilizes distinct policies for each task. The operational range and desired gaits are tuned for each specific task, with minor fine-tuning of standard reward coefficients or parameter $\delta$ to optimize performance. Results are detailed in the \href{https://www.youtube.com/watch?v=JV2_HfTlOKI}{supplementary videos}.

\textbf{\href{https://www.youtube.com/watch?v=JV2_HfTlOKI&t=54}{Traversing Outdoor Terrain}} \enspace The quadruped locomotion policy was learned in a simulation with bumpy terrain, slopes (up to 27$^\circ$), stairs (up to 20\,cm), and steps (up to 34.5\,cm). The experiments were conducted on the KAIST campus and nearby mountains. The robot successfully traversed slippery hills, rocks, steep slopes, stairs, and deformable terrain.

\textbf{\href{https://www.youtube.com/watch?v=JV2_HfTlOKI&t=82}{Climbing High Step}} \enspace Our method demonstrated the ability to overcome 58\,cm and 67\,cm obstacles in hardware experiments using HOUND and HOUND2, respectively. To verify robustness, we conducted five tests for each case, all of which were successful. To our knowledge, this achievement represents the highest obstacle clearance in blind locomotion of a quadruped robot. 
Considering larger robots' advantage in overcoming higher steps, we defined a \textit{Step-to-Leg} ratio by dividing the maximum step height by the combined leg lengths (thigh and calf) for fair comparisons. 
Lee et al.~\cite{lee2020learning}, Nahrendra et al.~\cite{nahrendra2023dreamwaq}, and Wu et al.~\cite{wu2023learning} achieved step heights of 16.8\,cm (\textit{Step-to-Leg} 0.27) with the ANYmal, 15\,cm (\textit{Step-to-Leg} 0.38) with the Unitree A1, and 25\,cm (\textit{Step-to-Leg} 0.59) with the Unitree Go1, respectively. Our results for both absolute step height (67\,cm) and the \textit{Step-to-Leg} ratio (0.96 for HOUND2) outperform those of existing RL-based blind locomotion studies on quadruped robots. 

\begin{figure} [ht!]
    \begin{center}
        \includegraphics[width=1.0\columnwidth]{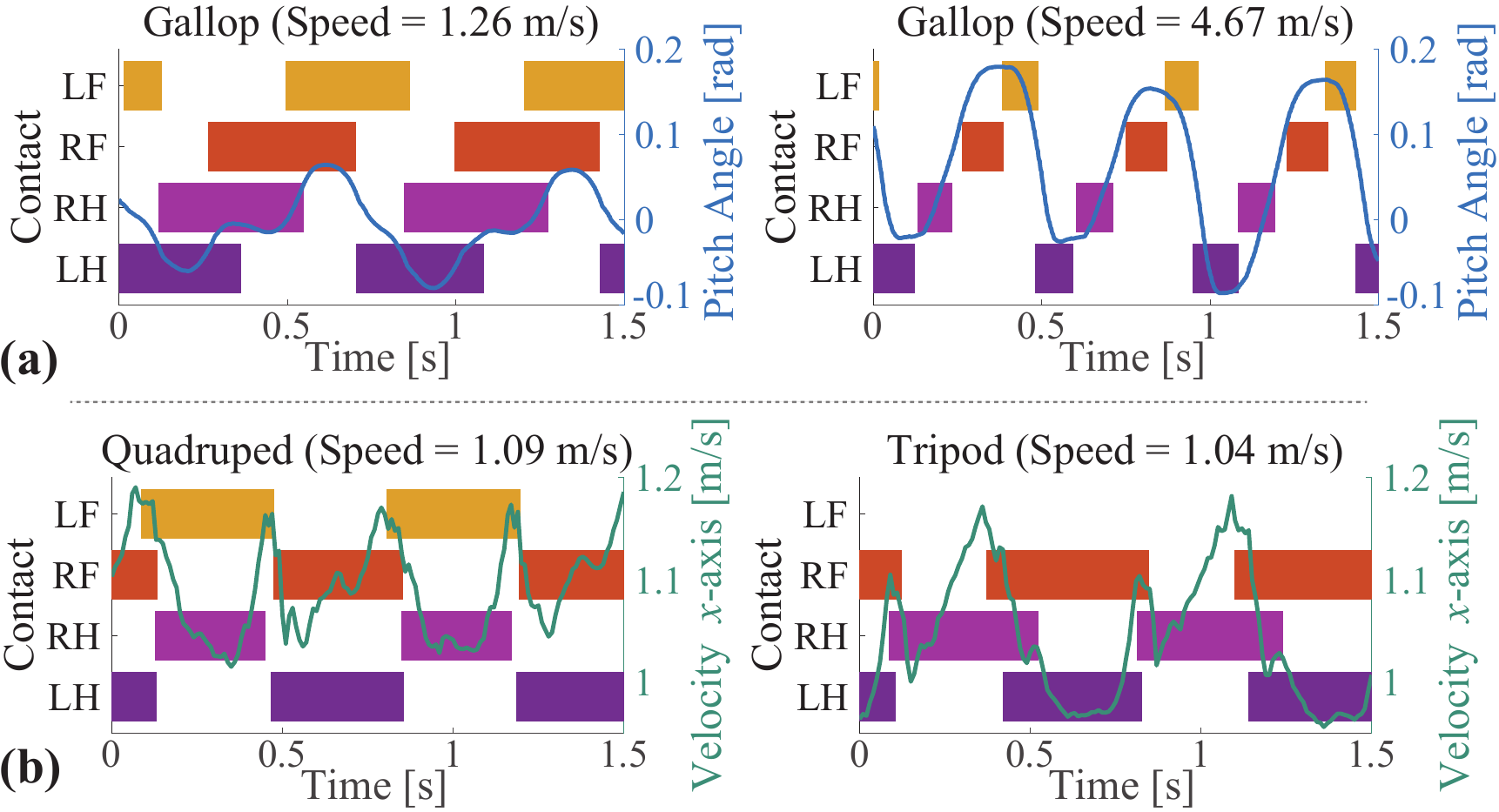}
    \end{center}
    \caption{Experimental results of gait adjustments: (a) Galloping at forward velocities of 1.26\,m/s (left) and 4.67\,m/s (right), with estimated contact states and pitch angles. (b) Quadruped trotting (left) and tripod mode locomotion (right), with estimated contact states and \textit{x}-axis velocities.}
    \vspace{-15pt}
    \label{figure:gallop}
\end{figure}

\textbf{\href{https://www.youtube.com/watch?v=JV2_HfTlOKI&t=101}{Running with Gallop Gait}} \enspace We demonstrated agile running of a quadruped with a galloping gait on a treadmill. Fig.~\ref{figure:gallop} (a) shows that the gait and pitch motions are adjusted based on running speeds of 1.26\,m/s (left) and 4.67\,m/s (right). Although the reference stance time $T_{\text{st}}$ is heuristically derived (stride length over command velocity), the gait constraint \( f_{i}\geq-0.3 \) allows for gait adjustments, facilitating agile running. Notably, the resultant stance time of 0.40\,s for 1.26\,m/s is 79\,\% of the desirable range \([0.29, 0.43]\)\,s, and 0.11\,s for 4.67\,m/s is 11\% of the desirable range \([0.10, 0.19]\)\,s, where both ranges are defined by \( f_{i}\geq-0.3 \).

\textbf{\href{https://www.youtube.com/watch?v=JV2_HfTlOKI&t=119}{Tripod Mode}} \enspace The robot can maintain one leg lifted (LF leg in this experiment), induced by a barrier reward, enabling tripod walking. The robot can walk, climb stairs with a tripod trot gait, and smoothly transition to a quadruped policy during locomotion. Fig.~\ref{figure:gallop} (b) shows the gait and \textit{x}-axis velocity for quadruped and tripod modes. The preferred gait is set to a trot, but the gait constraint \( f_{i} \geq -0.6 \) allows for automatic adjustments in tripod motion. The RF and LH legs initiate contact before the RH contact ends to compensate for the body’s drop due to the absence of LF contact. This results in extended RF and LH stance time of 0.46\,s, which is 86\% of the desirable stance time range \([0.22, 0.50]\,\text{s}\) (\( f_{i} \geq -0.6 \)).

\textbf{\href{https://www.youtube.com/watch?v=JV2_HfTlOKI&t=145}{Biped Mode}} \enspace We demonstrated that the quadruped robot HOUND can walk and run in biped mode, following commanded velocities. Even with point feet, the robot resists hard pushes from both front and back by using the pitch motion of the hip joint. Notably, it also withstands lateral pushes, even without an actuated DOF for roll motion in its biped configuration. By using the same heuristics of varying \(T_{\text{st}}\) from galloping tasks, we achieved a bipedal locomotion speed of 3.6 m/s based on treadmill measurements. 
For the stair climbing task, the same stair environment used in quadruped mode is employed during learning. Finally, the robot performs biped locomotion while carrying a box. In the learning stage, the box's position and mass are added to the privileged state \(\mathbf{x}_t\), and the robot's nominal configuration is adjusted to support a box. An additional barrier reward ensures the box is properly positioned above the robot's arm. In real-world experiments, the robot carries a 7.5 kg load, demonstrating push recovery. 

\section{Conclusion}
In this study, we present a learning framework that enables various modes of operation (biped, tripod, quadruped) on a single quadruped robot, utilizing a barrier-based style reward. The framework effectively guides the desired motion style and facilitates adjustments to the preferred gait as required by the task. Extensive real-world experiments across diverse environments demonstrate effectiveness of the framework. Future work aims to generalize the barrier formulation to
develop a unified policy capable of handling diverse tasks.

\bibliographystyle{ref/IEEEtran}   

\end{document}